\documentclass{article}

\usepackage{PRIMEarxiv}

\usepackage[utf8]{inputenc} 
\usepackage[T1]{fontenc}    
\usepackage{hyperref}       
\usepackage{url}            
\usepackage{booktabs}       
\usepackage{amsfonts}       
\usepackage{nicefrac}       
\usepackage{microtype}      
\usepackage{lipsum}
\usepackage{fancyhdr}       
\usepackage{graphicx}       
\graphicspath{{media/}}     

\usepackage{amsmath}
\usepackage{amssymb}
\usepackage{multirow} 
\usepackage{longtable} 
\usepackage{array} 
\usepackage{makecell}
\usepackage{threeparttable} 
\usepackage{booktabs}

\title{Elastic Interaction Energy-Informed Real-Time Traffic Scene Perception
}

\author{
  Yaxin FENG$^{1}$, Yuan LAN$^{1}$, Luchan ZHANG$^{2*}$, Guoqing LIU$^{3*}$ and Yang XIANG$^{1,4}$\thanks{Corresponding author.} \\ 
$^{1}$Department of Mathematics, Hong Kong University of Science and Technology, \\ Clear Water Bay, Hong Kong SAR, China \\
$^{2}$College of Mathematics and Statistics, Shenzhen University, Shenzhen, China \\
$^{3}$Shenzhen Youjia Innov Tech Co., Ltd., Shenzhen, China\\
$^{4}$Algorithms of Machine Learning and Autonomous Driving Research Lab, HKUST Shenzhen-Hong Kong \\ Collaborative Innovation Research Institute, Futian, Shenzhen, China \\
\texttt \small{\{yfengba, ylanaa\}@connect.ust.hk}, \texttt \small{zhanglc@szu.edu.cn}, \texttt \small{guoqing@minieye.cc}, \texttt \small{maxiang@ust.hk} \\
}

\begin{document}
\maketitle

\begin{abstract}
Urban segmentation and lane detection are two important tasks for traffic scene perception. Accuracy and fast inference speed of visual perception are crucial for autonomous driving safety. Fine and complex geometric objects are the most challenging but important recognition targets in traffic scene, such as pedestrians, traffic signs and lanes. In this paper, a simple and efficient topology-aware energy loss function-based network training strategy named EIEGSeg is proposed. EIEGSeg is designed for multi-class segmentation on real-time traffic scene perception. To be specific, the convolutional neural network (CNN) extracts image features and produces multiple outputs, and the elastic interaction energy loss function (EIEL) drives the predictions moving toward the ground truth until they are completely overlapped. Our strategy performs well especially on fine-scale structure, \textit{i.e.} small or irregularly shaped objects can be identified more accurately, and discontinuity issues on slender objects can be improved. We quantitatively and qualitatively analyze our method on three traffic datasets, including urban scene segmentation data Cityscapes and lane detection data TuSimple and CULane. Our results demonstrate that EIEGSeg consistently improves the performance, especially on real-time, lightweight networks that are better suited for autonomous driving.
\end{abstract}


\section{INTRODUCTION}
Semantic segmentation is a pixel-level classification of images, which is a dense perception task. Among them, the low-latency, high-efficiency segmentation model is widely used in the autonomous driving industry that requires real-time reasoning or limited computing resources, such as advanced driver assistance systems (ADAS)~\cite{pan2018spatial}, drones~\cite{tan2021flying}, and mobile robots. In this work, we mainly focus on two traffic scene perception tasks for autonomous vehicles, \textit{i.e.} urban scene segmentation and segmentation-based lane detection. In the urban scene, accurate multi-class semantic segmentation is required to distinguish the drivable areas (roads) from sidewalks, buildings, vehicles, pedestrians, traffic signs and so on~\cite{romera2017erfnet,yu2021bisenet}. In the lane detection problem, the existence and location of different lane instances need to be inferenced in real time, even when they are occluded by cars and crowd~\cite{pan2018spatial,hou2019learning}.


At present, vast studies of visual perception focus on the improvement of neural network structure, such as deeper and denser networks~\cite{yuan2020object,wang2020deep,kirillov2023segment} and attention-based methods such as vision-transformer~\cite{han2022survey, dosovitskiy2020image}. However, although complex network structures with huge number of parameters promise sophisticated performance, they will bring significant computational cost and have relatively slow inference speed. Thus these networks are not practical for many autonomous driving assistant systems that require real-time perception under limit computing resources~\cite{son2023lightweight}. Consequently, the study on real-time traffic scene perception on low compute devices are necessary. 

\begin{figure}[ht]
\centering
\includegraphics[width=0.8\textwidth]{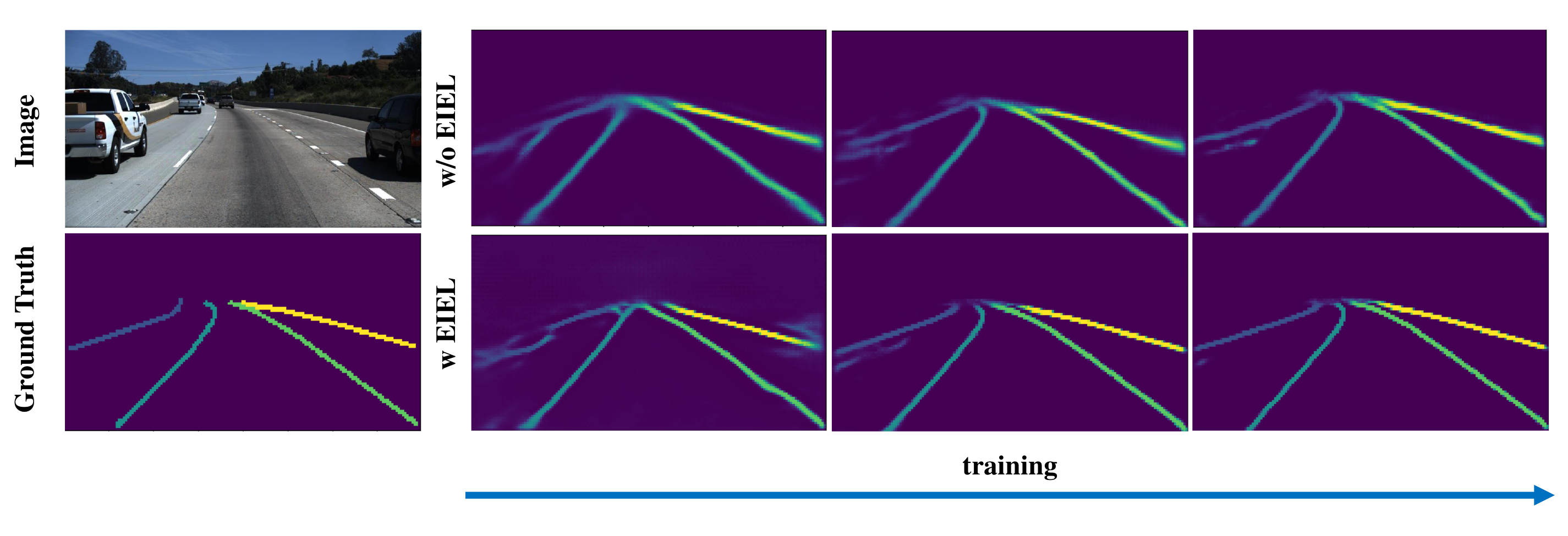} 
\caption{An example of training process. The pictures from left to right are the predictions of the models without/with EIEL during training. EIEL corrected the wrong classification pixels, and the end of the lane is clearer.}
\label{fig.process}
\end{figure}

On the other hand, because the prediction errors often appear on geometry complex objects such as disconnected at/missing slender structures with only a few pixels inside, and blurry on boundaries of irregular shapes, another prevailing direction is to develop tolopogy-aware approaches. The main branch of the boundary refinement methods is post processing~\cite{blake2011markov,lafferty2001conditional,yuan2020segfix}, which will slow the inference speed. Another type of methods improve boundary segmentation during training, \textit{e.g.} developing boundary-aware loss functions~\cite{ke2018adaptive,ding2019boundary,borse2021inverseform,wang2022active,kervadec2019boundary,takikawa2019gated, lan2020elastic}. Our approach belongs to the second category, which utilizes a global topological-aware loss function to guide lightweight neural networks to extract more image features and global contextual information during the training phase, resulting in better performance compared to the baseline; see Fig.~\ref{fig.process}. We believe that our approach will also be effective in other visual perception tasks, and larger size models on other application such as high resolution image segmentation~\cite{shen2022high}.

\begin{figure*}[htbp]
\centering
\includegraphics[width=1.0\textwidth]{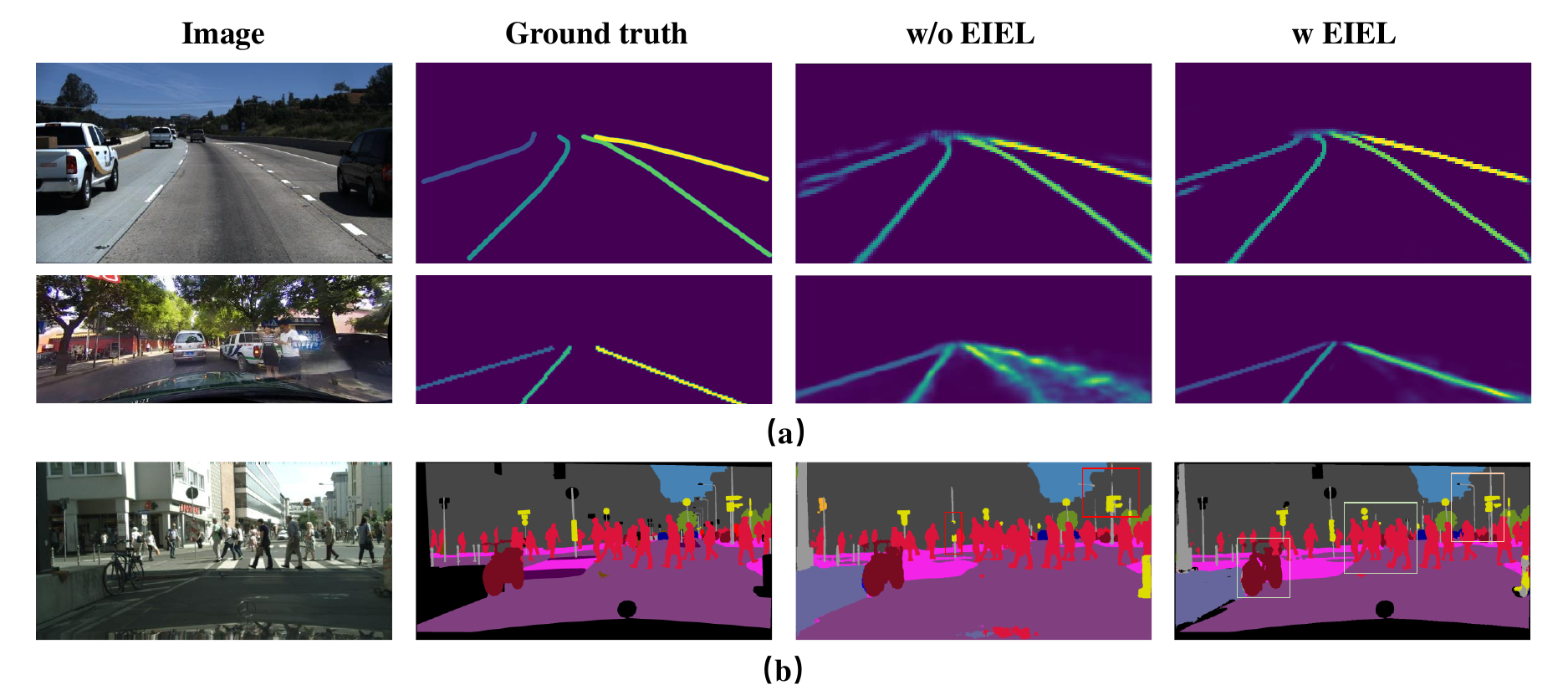} 
\caption{Comparison of the output maps between Baseline ERFNet ~\cite{romera2017erfnet} without (w/o) and with (w) EIEL in (a) lane detection and (b) segmentation. To be specific, the pictures from left to right are: raw image, ground truth, outputs w/o EIEL and outputs w EIEL.}
\label{fig1}
\end{figure*}

We propose a topology-aware loss-informed training strategy for multi-class semantic segmentation, named EIEGSeg (Elastic Interaction Energy Guided Segmentation). It is a plugged-in training strategy in the sense that our EIEL can be integrated in any network structures to improve the segmentation effect of complex geometries under class imbalance, without increasing computational complexity and inference time. Our loss function is inspired by the line defects of dislocation in crystals~\cite{hirth1983theory}. In image segmentation problems, this energy considers pixels of an object as a whole, rather than considering distance between pairwise pixels on the boundary, thus long-thin objects tend to be connected as the energy descents, and the edges of complex topological shapes are clearly outlined. Figure \ref{fig1} shows some effective applications by adding EIEL on a baseline. Slender objects such as poles and occluded lane lines are better connected and identified, boundaries of small and irregular shape objects such as traffic sign, traffic light, pedestrians, riders, etc. can be clearer recognized. More details of the method will be provided in Sec.~\ref{sec:EIE}. 

We evaluate our method on three popular autonomous driving scene benchmarks, \textit{i.e.} one urban scene data Cityscapes~\cite{cordts2016cityscapes} and two lane detection datasets Tusimple~\cite{tusimple2017benchmark} and CULane~\cite{pan2018spatial}. We compare the performance based on a real-time network structure ERFNet~\cite{romera2017erfnet}, BiSeNetV1~\cite{yu2018bisenet} with or without EIEL function. A popular non-real time FCN semantic segmentation baseline HRNet-W48-OCR (OCR)~\cite{yuan2020object} is also fine-tuned with EIEGSeg strategy. The experiment results show that training with EIEL can consistently improve the mIoU for segmentation, accuracy, F1-score for lane detection based on different network backbones. 

The main contributions are as follow:

\begin{itemize}
\item We propose a simple and efficient topology-aware energy-based loss function EIEL for multi class segmentation. Small, slender and irregularly shaped objects can be identified more accurately, and discontinuity issues on slender objects can be significantly improved comparing with the baselines.
\item The EIEGSeg training strategy is independent of network selection, and can be applied to any end-to-end network structure during training stage. Our approach works well on both lightweight and high-resolution networks, with prominent improvements in lightweight networks that are suitable for assisting autonomous vehicles in real-time inference.
\item Three datasets are used to evaluate our method. In Cityscapes, the prediction accuracy on fine-scale structure objects boosts remarkably. In TuSimple and CULane, the lane segmentation is much clearer and the disconnection problem caused by occlusion is fixed. 
\end{itemize}

\section{RELATED WORK}
\subsection{Semantic Segmentation Network Structures}
The fast development of end-to-end deep learning-based segmentation started from fully convolutional network (FCN) ~\cite{long2015fully}. Many following works are based on this encoder-decoder network structure. In addition to those non-real time networks that have deeper and denser layers~\cite{yurtkulu2019semantic,zhou2019unet++, wang2020deep,kirillov2023segment} or transformer series attention-based network~\cite{dosovitskiy2020image}, more and more studies focus on design lightweight network delicately~\cite{romera2017erfnet,yu2018bisenet,yu2021bisenet,wang2019lednet,dong2020real} and can achieve real-time performance that balances the effectiveness and the efficiency. In the lane segmentation problem, SCNN \cite{pan2018spatial} proposes the spatial information passing network structure blocks, which can also be used in semantic segmentation.

\subsection{Topology-aware segmentation}
There are two branches of segmentation refinement, i.e., post-processing and applying topology-aware technique. 

Post-refinement such as markov random fields (MRF) ~\cite{blake2011markov} and conditional random fields (CRFs) ~\cite{lafferty2001conditional} refine the outputs of network. Clustering is utilized to distinguish the instances type in lane detection task after training~\cite{abualsaud2021laneaf}. Segfix ~\cite{yuan2020segfix} can transfer the interior pixels to be boundaries by training a separate network. However, the end-to-end network architecture without post-process performs better for real-time inference in most cases.

Some works on segmentation geometry refinement focus on object boundary. Pairwise pixel-level affinity fields have been used to classify different pixels, \textit{e.g.} ~\cite{ke2018adaptive,ding2019boundary,kervadec2019boundary}. InverseForm ~\cite{borse2021inverseform} designs an inverse-transformation network to learn the boundaries distance. Loss functions in~\cite{takikawa2019gated} are studied and proposed to refine the boundary details. Most of these training processes are combined with cross-entropy (CE) and fuse different layer features to improve the details. However, they are not examined on lightweight network architectures, thus remaining a question that whether they can significantly improve real-time networks. Also, most of these methods are not sensitive in long-thin or fine-scale objects. 

Others methods focus on topology relation of pixels or refine the class imbalance problem. Intersection over union (IoU) is transformed into differentiable form loss function to guide the refinement of pixels prediction ~\cite{berman2018lovasz}. ~\cite{lan2020elastic,xiang2006active} utilized elastic interaction energy loss function to achieve connection improvement on medical binary segmentation. ~\cite{hu2019topology} uses a topology-preserving loss function based on the thought of level set method and Betti number in conjunction with binary cross energy to improve the connectivity of slender objects, but it was only applied in binary problems, \textit{e.g.} eye blood vessels, remote sensing data and road crack segmentation.

\begin{figure*}[htbp]
\centering
\includegraphics[width=1.0\textwidth]{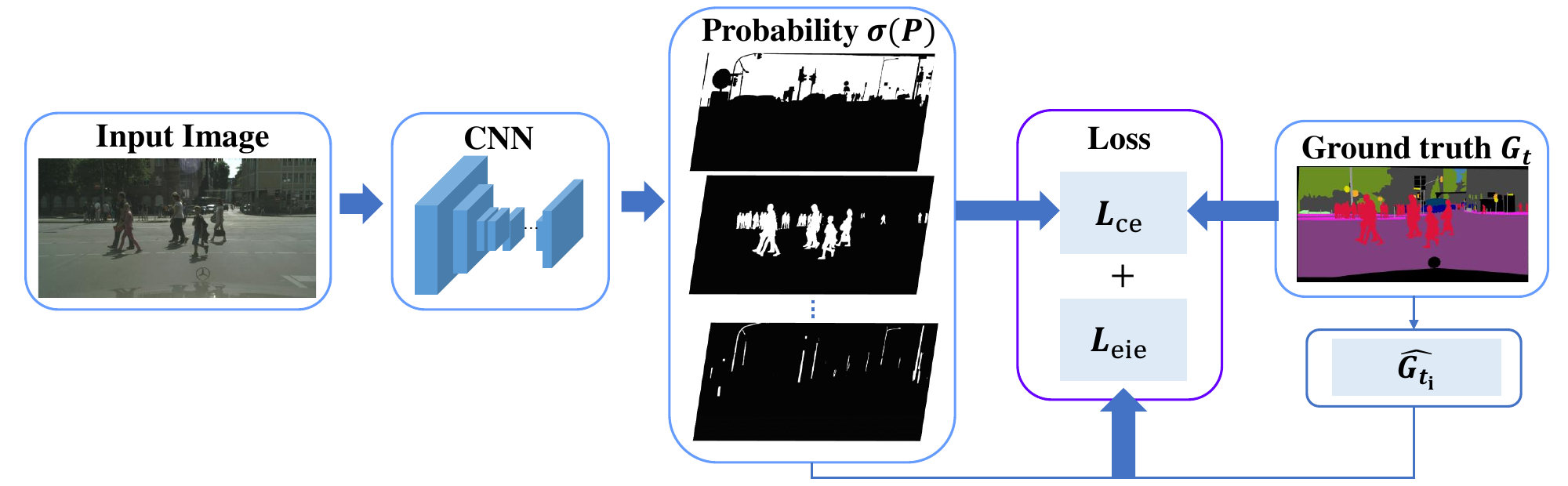} 
\caption{Network Structure of EIEGSeg, which jointly guided by EIEL and CE in the training process. Note that this is just a schematic illustration, and the predicted probability maps of some complicated cases in the figure do not exactly correspond to reality.}
\label{fig2}
\end{figure*}

\section{METHODOLOGY}
\subsection{Network Structure}
The proposed strategy EIEGSeg is jointly trained by the guidance of EIEL and CE. The overall architecture of EIEGSeg is shown in figure \ref{fig2}. EIEL and CE are jointly applied in the training process, while the CE help classifying different categories of pixels and the EIEL refines the topology of each class. Since EIEGSeg is a multi-class segmentation based strategy, the network predicts N output maps for N-class objects and the ground truths are adjusted into the one-hot form when applying EIEL. The overall training strategies of our two visual perception tasks are similar, \textit{i.e.} urban scene segmentation and lane detection, except that a post process is required in lane detection tasks, in which lane coordinates are collected from the predicted instance masks. 

\subsection{Elastic Interaction Energy}\label{sec:EIE}
\subsubsection{Preliminary}

The elastic interaction energy functional defined in~\cite{hirth1983theory} is
\begin{equation}\label{eq:eie}
E=\int_{\gamma} \int_{\gamma^{\prime}} \frac{d \boldsymbol{l} \cdot d \boldsymbol{l}^{\prime}}{r} 
\end{equation}
In the above functional, $\gamma^{\prime}$ is the curves with other parameter $s^{\prime}$, vector $d \boldsymbol{l}$ represents line element on one set of curves $\gamma$ with tangent direction $\boldsymbol{\tau}$, i.e. $d\boldsymbol{l}=\boldsymbol{\tau} d l$, and the $r$ is Euclidean distance between (x,y) and ($\left.\mathrm{x}^{\prime}, \mathrm{y}^{\prime}\right)$, i.e. $r=$ $\sqrt{\left(x-x^{\prime}\right)^2+\left(y-y^{\prime}\right)^2}$. 

\subsubsection{EIEL for Image Segmentation}

In our tasks, we define the prediction $\gamma_p$ and ground truth $\gamma_{gt}$ between two sets of curves, and Eq.~\ref{eq:eie} becomes:
\begin{equation}\label{eq:eie2}
\begin{aligned}
E & = E_s + E_i\\
  & =\frac{1}{8 \pi} \int_{\gamma_{gt} \cup \gamma_p} \int_{\gamma_{gt}^{\prime} \cup \gamma_p^{\prime}} \frac{d \boldsymbol{l} \cdot d \boldsymbol{l}^{\prime}}{r} \\
  & =\frac{1}{8 \pi} \int_{\gamma_{gt}} \int_{\gamma_{gt}^{\prime}} \frac{d \boldsymbol{l}_{\mathbf{gt}} \cdot d \boldsymbol{l}_{\mathbf{gt}}^{\prime}}{r}+\frac{1}{8 \pi} \int_{\gamma_p} \int_{\gamma_p^{\prime}} \frac{d \boldsymbol{l}_{\mathbf{p}} \cdot d \boldsymbol{l}_{\mathbf{p}}^{\prime}}{r} \\
  & \quad +\frac{1}{4 \pi} \int_{\gamma_{gt}} \int_{\gamma_{p}} \frac{d \boldsymbol{l}_{\mathbf{gt}} \cdot d \boldsymbol{l}_{\mathbf{p}}}{r},
\end{aligned}
\end{equation}
where the vector $d\boldsymbol{l}$ is the sum of ground truth $\boldsymbol{l_{gt}}$ and the predicted $\boldsymbol{l_p}$, and their directions are opposite, $\boldsymbol{dl}=\boldsymbol{dl_{gt}}+\boldsymbol{dl_p}$.

\begin{figure}[ht]
\centering
\includegraphics[width=0.5\textwidth]{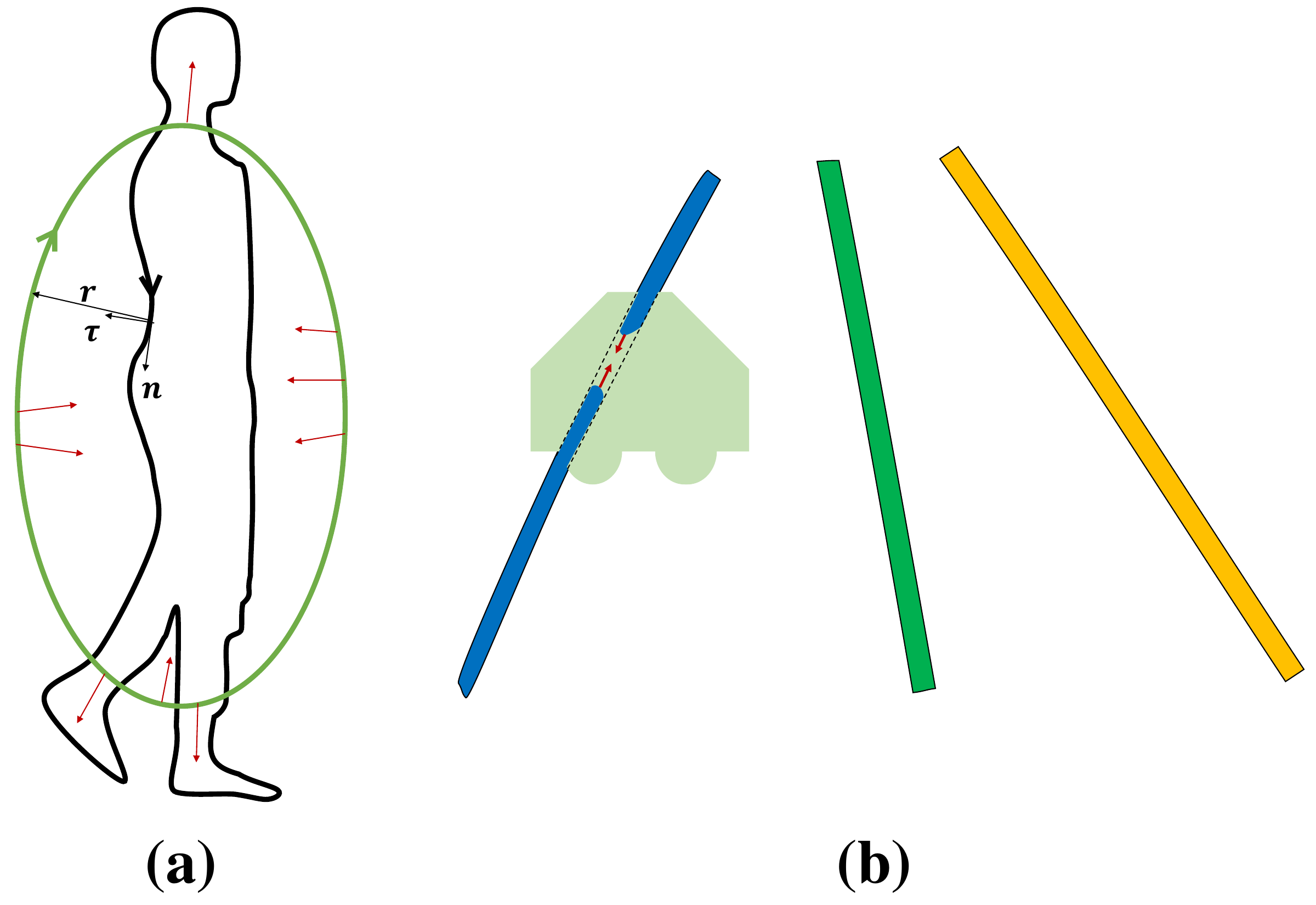} 
\caption{Description of how the EIEL works. (a) shows how the moving curve being attracted to the ground truth. (b) shows how the disconnected prediction of the left most lane is corrected by EIEL even it is occluded by a car.}
\label{fig.peopleandlane}
\end{figure}

In Eq.~\ref{eq:eie2}, the self-energy $E_s$ consists of the first two terms for the self-energies of the prediction curve set $\gamma_p$ and the ground truth curve set $\gamma_{gt}$, while the third term is the interaction energy $E_i$ between the two sets of curves  $\gamma_p$ and $\gamma_{gt}$.
Since the prediction curve $\gamma_p$ and the ground truth curve $\gamma_{gt}$ have opposite orientations, the elastic interaction between them is strongly attractive based on the third term in Eq.~\ref{eq:eie2}, and the prediction curve $\gamma_p$ is attracted to the  ground truth curve $\gamma_{gt}$ to minimize the total energy in Eq.~\ref{eq:eie2}; see Fig.\ref{fig.peopleandlane} (a). When the two curves coincide, the energy minimum state is reached, and the ground truth is identified perfectly by the prediction. Under this elastic energy, the boundaries of the disconnected parts of the prediction are also attracted to each other for the same reason, and eventually combine; see Fig.\ref{fig.peopleandlane} (b). Besides, the self-energy of the prediction tends to make the prediction curves smoothing, because a longer curve has a larger self-energy. More mathematical properties of this elastic loss can be found in ~\cite{lan2020elastic,xiang2006active}.


In the image space, the boundary of ground truth $G_t$ is expressed as $\nabla G_t = \nabla (G_t-0.5) = -\delta (\gamma_{gt})\boldsymbol{n}_{gt}$. If the predicted probability of the moving predicted curves is $p$, the curves can be expressed implicitly by gradient of $\Phi_p = 0.5-p$ ranging in $[-0.5,0.5]$, \textit{i.e.} $\nabla \Phi_p = -\delta (\gamma_{p})\boldsymbol{n}_{p}$. Note that $\boldsymbol{n}_{gt} \text{ and } \boldsymbol{n}_{p}$ are unit normal vectors of curves (boundaries).


Thus, the elastic interaction energy functional as well as the EIE loss should be:
\begin{equation}\label{eq.leie}
\begin{split}
&E(G_t,\Phi_p) =\frac{1}{8 \pi} \int_{\mathbf{R}^2} d x d y \\
&\int_{\mathbf{R}^2} \frac{\nabla\left(G_t+\alpha \Phi_p \right)(x, y) \cdot \nabla\left(G_t+\alpha \Phi_p \right)\left(x^{\prime}, y^{\prime}\right)}{r} d x^{\prime} d y^{\prime} 
\end{split}
\end{equation}

In order to apply this loss function in neural network, the velocity field as well as the gradient descent direction is required in back propagation. After applying variation of the functional Eq.~\ref{eq.leie} about the prediction $\Phi_p$, the velocity field of prediction obtained:
\begin{equation}
   v(\Phi_p)=-\frac{\delta E}{\delta \Phi_p}= \frac{1}{4 \pi} \int_{\Omega} \frac{\boldsymbol{r}}{r^3} \cdot \nabla (G_t +\alpha \Phi_p)\left(x^{\prime},y^{\prime} \right)  d x^{\prime}d y^{\prime}
\end{equation}

\subsubsection{EIEL-Guided Multi Class Segmentation}

In multi class learning, $N$ feature maps are output at the last layer of neural network. Each output map is represented by $P_i$ and Softmax $\sigma$ is applied to convert them into probability distribution of each pixel, i.e. $\sigma: P \rightarrow p \in [0,1]^{N*h*w}$. Then the implicit function of predictions is $\Phi_i = 0.5-\sigma(P)_i$,
and the corresponding expression of ground truth with opposite orientation is $\hat{{G_t}_i}-0.5$, where $\hat{{G_t}_i}$ is the one-hot mode of the ground truth, $\hat{G}_{t_i}=\mathbf{1}_{G_{t_i}}.$

According to Eq.~\ref{eq.leie}, the EIEL in multi-class segmentation is:
\begin{equation}
L_{\text {eie }}=\sum_{i}E(\sigma(P)_i,\hat{{G_t}_i}).
\end{equation}

In $N$ class semantic segmentation, each pixel on an image has a class label from 0 to $N-1$, within which the background is class 0 (the background is classified as ignore-label 255 in segmentation task, thus some unpredicted areas remain; see the last two columns in Fig~\ref{fig.seg}). Therefore, the output layer has N feature prediction maps, i.e. the output size is $B*N*h*w$.

\subsubsection{Efficient Calculation}
As the energy function $L_{eie}$ in Eq. \ref{eq.leie} is an integral over the whole image domain, fast Fourier transform can make the computation more efficient, by which the computation cost is reduced from $O(N^2)$ to $O(N\log N)$.
Assume the Fourier transform of $G_t+\alpha \Phi_p$ is $d_{mn}$, where $m$ and $n$ are frequencies in the Fourier space. 
The Fourier transform of $\frac{1}{R}=\frac{1}{\sqrt{x^2+y^2}} \text { is } \frac{\widehat{1}}{R}=\frac{1}{2 \pi \sqrt{m^2+n^2}}.
$
Therefore, in the Fourier space, the EIEL can be transformed into the form as
\begin{equation}
L_{\text {eie }}=\sum_{m, n} \sqrt{m^2+n^2} \cdot\left|d_{m n}\right|^2,
\end{equation}
and the gradient of $L_{eie}$ is the inverse Fourier transform, which is used in backward propagation
\begin{equation}
\frac{\delta L_{\text {eie }}}{\delta \Phi_p}=\mathcal{F}^{-1}\left(\frac{\sqrt{m^2+n^2}}{2} d_{m n}\right).
\end{equation}

\subsubsection{Total loss function in EIEGSeg}

The training loss $L_t$ in EIEGSeg is 
\begin{equation}
L_t=\lambda_1 L_{\text {eie }}+\lambda_2 L_{\text {ce }}\left(P, G_t\right).
\end{equation}
where $L_{\text {eie }}$ represents EIE loss and $L_{\text {ce }}$ is the cross-entropy loss. $\lambda_i, i=1,2$ are parameters. 


\section{EXPERIMENTS}
\subsection{Implimentation Details}
In semantic segmentation problem, we upsample the feature size and get end-to-end pixelwise prediction. The size of input images are H and W, and the output is with the size of $B*C*h*w$, where $C$ is the number of output feature map, $h=H/s$, $w=W/s$.

In lane detection problems, we upsample the features into $1/4$ of the original size, i.e. $B*C*h*w$ where $s=4$, because lanes are well structured, and do not need so many pixels to represent them. The ground truth is also down-scale to $1/4$ with linear interpolation. After segmentation, we chose the pixels in the center each row (sampling equidistant on the $y-$direction) of the lane line as their coordinates.

The experiments are implemented on a GPU RTX 4000 and RTX 3060. 

\subsection{Datasets and Evaluation Metrics}
\subsubsection{Cityscapes.}
In Cityscapes~\cite{cordts2016cityscapes}, dense annotations with 19 object classes are provided. The training and validation data amount is 2975 and 500, and we use the validation data in testing. We apply the official metric mean Intesection-over-Union (mIoU) and the IoU of each class to measure the performance overall and in specific categories.

\subsubsection{TuSimple.}
Tusimple~\cite{tusimple2017benchmark} has a maximum of five lane lines on highway and most of which are in daytime and clear weather, which includes 3626 training data and 2782 testing data. The official metric of the TuSimple dataset is accuracy (Acc), the false positive (FP) and the false negative (FN) are evaluated. In addition, we also compute the F1 score, and the average pixel-wise F1 score (Pix. F1) of the predictions. The Acc defined in TuSimple benchmark is: $Acc=\frac{N_{p r e d}}{N_{g t}},$ where $N_{p r e d}$ is the number of correct predicted lane points and $N_{g t}$ is the amount of ground-truth lane points.

\subsubsection{CULane.}
CULane~\cite{pan2018spatial} includes eight difficult-to-detect urban environments such as crowded, darkness, no-line, and shadows, and marks up to four lane lines. Data volume is 88880/9675/34680 as train/val/test. The official metric for CULane is F1 score based on IoU: $F 1=2 \cdot\left(\frac{P  \cdot  R }{P + R}\right),$
where $P$ defined as $\frac{TP}{TP+FP}$, $R$ defined as  $\frac{TP}{TP+FN}$. $TP$ is the correctly predicted lane points, $FP$ is the false positives number, and $FN$ is false negatives ones.

\begin{figure*}[htbp]
\centering
\includegraphics[width=1.0\textwidth]{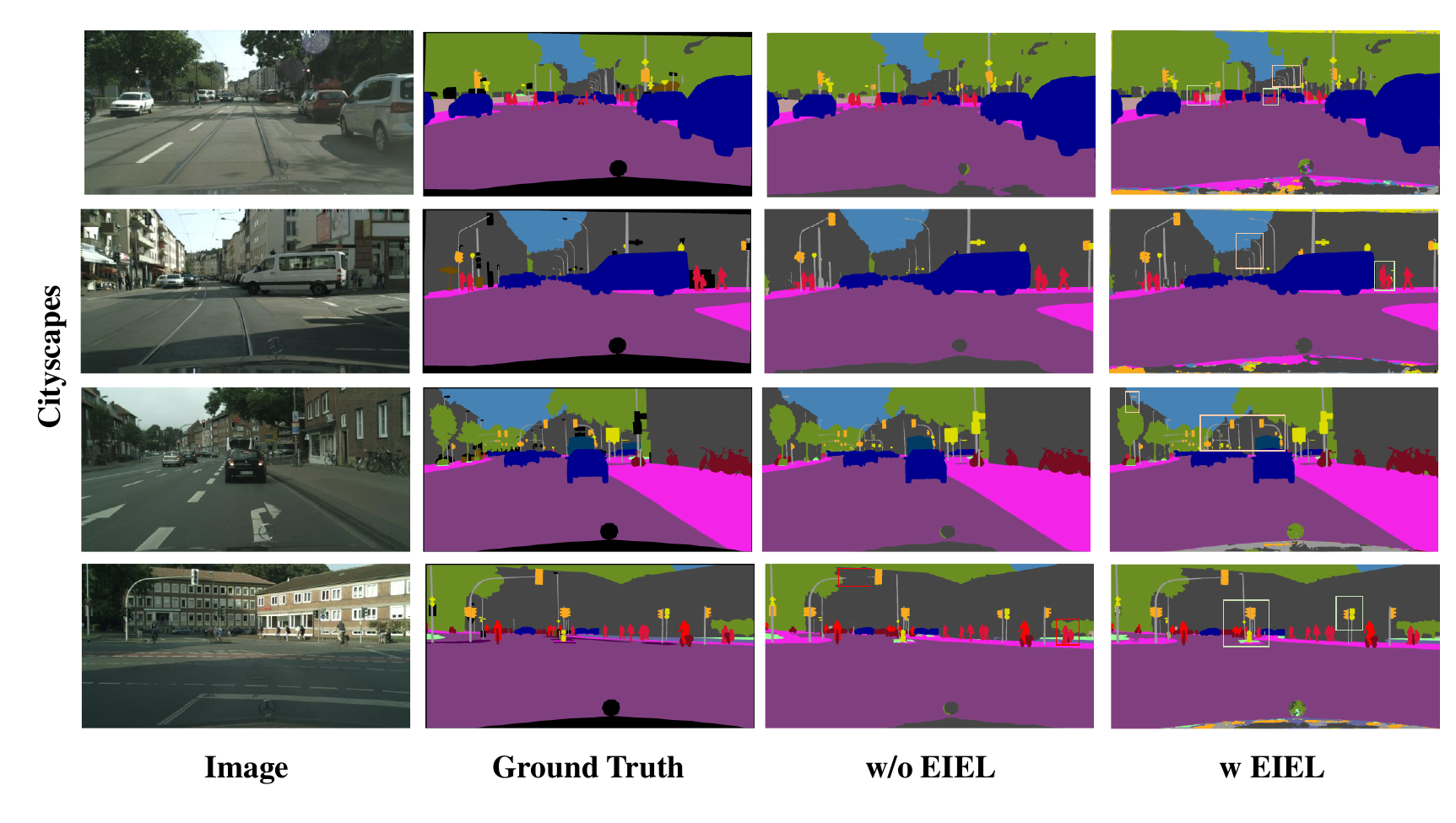} 
\caption{Comparison of the performance on Semantic Segmentation task. The pictures above from left to right are ground truth, label, baseline OCR without (w/o) and with (w) EIEL. Orange boxes indicate the better performance on pole, traffic sign and light, and the green ones are those with clearer pedestrians segmentation. Some obvious incorrect segmentation positions (w/o EIEL) in the third column are highlighted in red boxes.}
\label{fig.seg}
\end{figure*}

\begin{figure*}[ht]
\centering
\includegraphics[width=1.0\textwidth]{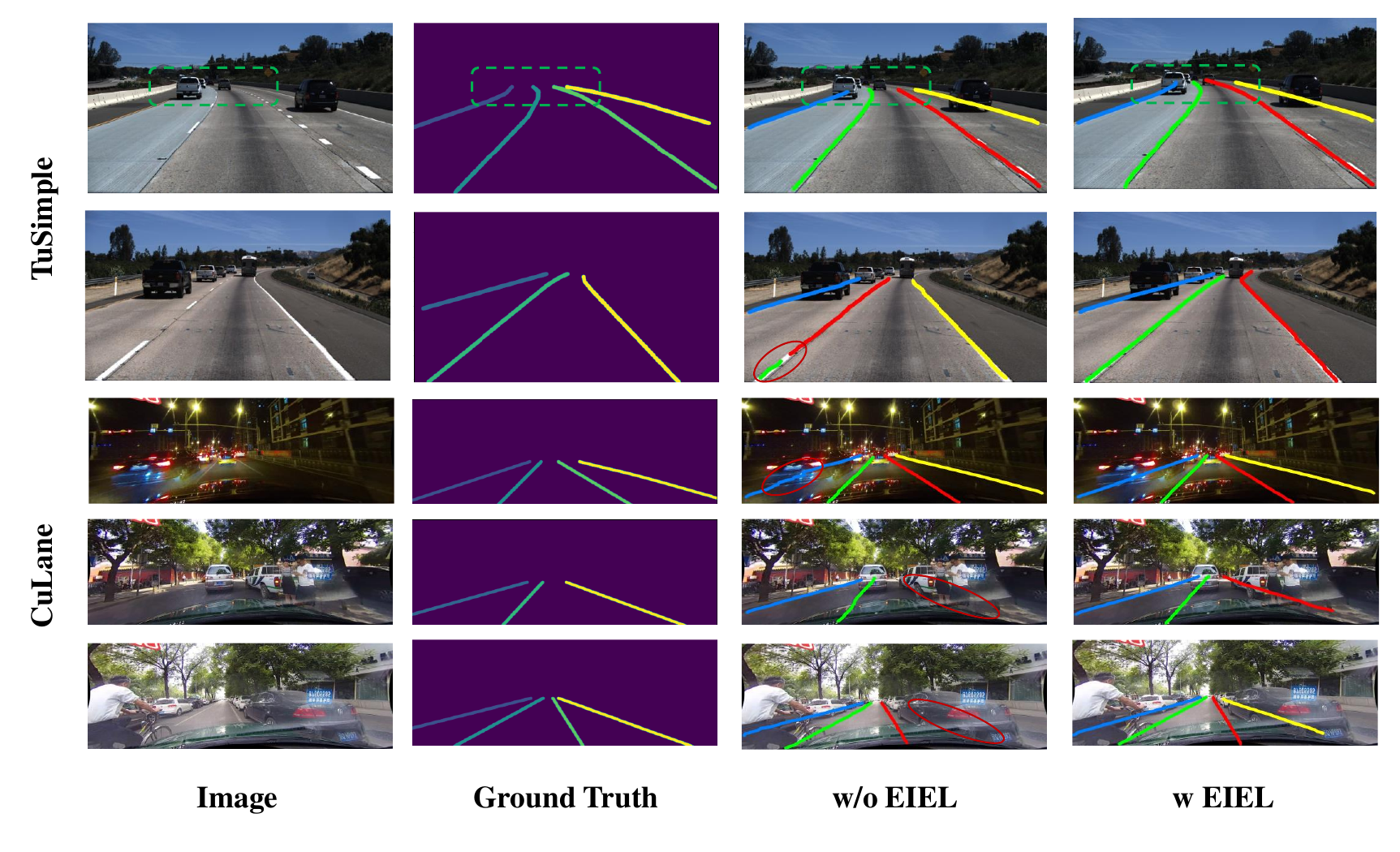} 
\caption{Performance on Lane Detection task with backbone ERFNet. From top to bottom, the baseline without EIEL are not curvy enough, line disconnected caused by wrong classify, bend as feature map blurry, lane missing caused by occlusion.}
\label{fig.lane}
\end{figure*}

\subsection{Results}
\subsubsection{Semantic segmentation.}
Table \ref{tab.city.realtime} shows performance on Cityscapes validation dataset. As shown in Tab.\ref{tab.city.realtime}, the predictions of wall, pole, traffic light, traffic sign, pedestrian, rider, bicycle are improved consistently when using different backbones with EIEL.

\begin{table*}[ht]\tiny
\caption{Class-wise mIoU results on Cityscapes validation set. First seven rows are real-time approaches, and the last seven rows are non real-time ones. $^\dag$ are our implementation. $^*$ means the results of multi-scale inference.}\label{tab.city.realtime}
\begin{center}
\setlength{\tabcolsep}{0.6mm}{
\begin{tabular*}{\hsize}{@{}@{\extracolsep{\fill}}l|llllllllllllllllllllll|l@{}}
\toprule
Method  &road  &\multicolumn{2}{c}{\makecell[c]{side\\ -walk}}  &bldg.  &wall  &fence  &pole   &\multicolumn{2}{c}{\makecell[c]{traffic\\light}}   &\multicolumn{2}{c}{\makecell[c]{traffic\\sign}} &veg.  &terrian  &sky  &person &rider &car &truck &bus &train &motor &bicycle &mIoU \\
\midrule
\midrule
ENet~\cite{paszke2016enet}  &96.3   &\multicolumn{2}{c}{\makecell[c]{74.2}}  &85.0 &32.2  &33.2  &43.5 &\multicolumn{2}{c}{\makecell[c]{34.1}}  &\multicolumn{2}{c}{\makecell[c]{44.0}}  &88.6  &61.4 &90.5  &65.5 & 38.4 & 90.6 &36.9 &50.5 &48.1 &38.8 &55.4 &58.3\\
LED~\cite{wang2019lednet}  &98.1   &\multicolumn{2}{c}{\makecell[c]{79.5}}  &91.6 &47.7  &49.9  &62.8  &\multicolumn{2}{c}{\makecell[c]{61.3}}  &\multicolumn{2}{c}{\makecell[c]{72.8}}  &92.6  &61.2  &94.9  &76.2 & 53.7 & 90.9 &64.4 &64.0 &52.7 &44.4 &71.6 &70.6\\
LBNAA~\cite{dong2020real}  &98.2   &\multicolumn{2}{c}{\makecell[c]{84.0}}  &91.6 &50.7  &49.5  &60.9  &\multicolumn{2}{c}{\makecell[c]{69.0}}  &\multicolumn{2}{c}{\makecell[c]{73.6}}  &92.6  &70.3  &94.4 &83.0 & 65.7 & 94.9 &62.0 &70.9 &53.3 &62.5 &71.8 &70.65\\
\midrule  
ERF$^\dag$~\cite{romera2017erfnet}  &97.46   &\multicolumn{2}{c}{\makecell[c]{80.79}}  &90.28 &49.5  &55.6  &60.46  &\multicolumn{2}{c}{\makecell[c]{61.48}}  &\multicolumn{2}{c}{\makecell[c]{70.33}}  &91.07  &61.65  &93.3  &74.12 & 54.5 & 92.14 &62.24 &73.55 &60.34 &45.18 &68.44 &70.65\\

EIEGSeg(ERF)$^\dag$  &\bf{97.69}   &\multicolumn{2}{c}{\makecell[c]{80.46}}  &\bf{91.57} &\bf{58.48}  &53.19  &\bf{65.36}  &\multicolumn{2}{c}{\makecell[c]{\bf{66.86}}}  &\multicolumn{2}{c}{\makecell[c]{\bf{74.69}}}  &\bf{92.21}  &\bf{63.86}  &\bf{94.06}  &\bf{78.72} & \bf{59.76} & \bf{93.66} &\bf{71.08} &\bf{77.68} &\bf{64.93} &\bf{46.55} &\bf{72.1} &\bf{73.84}\\
\midrule  
BSV1$^\dag$~\cite{yu2018bisenet}  &97.93   &\multicolumn{2}{c}{\makecell[c]{83.42}}  &91.70 &47.21  &52.44  &63.13  &\multicolumn{2}{c}{\makecell[c]{70.41}}  &\multicolumn{2}{c}{\makecell[c]{77.76}}  &91.99  &61.64  &94.73  &80.64 & 59.60 & 94.50 &\bf{73.23} &\bf{83.25} &\bf{74.07} &\bf{58.34} &75.96 &75.37\\

EIEGSeg(BSV1)$^\dag$  &\bf{98.05}   &\multicolumn{2}{c}{\makecell[c]{\bf{84.45}}}  &\bf{92.42} &\bf{55.27}  &\bf{59.05}  &\bf{65.52}  &\multicolumn{2}{c}{\makecell[c]{\bf{70.97}}}  &\multicolumn{2}{c}{\makecell[c]{\bf{78.84}}}  &\bf{92.38}  &\bf{63.48}  &\bf{94.92}  &\bf{81.40} &\bf{61.80} &\bf{94.52} &66.47 &79.43 &72.82 &52.53 &\bf{76.43} &\bf{75.83}\\

\midrule
\midrule

LargeFOV~\cite{chen2017deeplab}
&98.3  &\multicolumn{2}{c}{\makecell[c]{78.8}} &89.9 &42.2 &52.2 &52.9 &\multicolumn{2}{c}{\makecell[c]{62.3}} &\multicolumn{2}{c}{\makecell[c]{71.1}} &91.0 &59.2 &92.2 &75.9 &52.1 &92.3 &48.8 &70.2 &37.6 &54.6 &72.3 &68.0\\

SCNN(R101)~\cite{pan2018spatial}
&98.3  &\multicolumn{2}{c}{\makecell[c]{86.1}} &92.6 &46.7 &61.2 &66.1 &\multicolumn{2}{c}{\makecell[c]{74.3}} &\multicolumn{2}{c}{\makecell[c]{81.5}} &92.7 &65.4 &94.7 &84.0 &65.5 &94.8 &57.7 &82.0 &59.9 &67.0 &80.1 &76.4\\

DeepLabV3~\cite{chen2017rethinking}  &98.4   &\multicolumn{2}{c}{\makecell[c]{86.5}}  &93.1 &\bf{63.9}  &62.6  &66.1  &\multicolumn{2}{c}{\makecell[c]{72.2}}  &\multicolumn{2}{c}{\makecell[c]{80.0}}  &92.8  &66.3  &95.0 & 83.3 & 65.5 &95.3 &74.5 &89.0 &80.0 &67.4 & 78.4 &79.5\\

\midrule
OCR~\cite{yuan2020object}  &\bf{98.52}  &\multicolumn{2}{c}{\makecell[c]{\bf{87.5}}} &\bf{93.67} &62.15 &\bf{66.47} &71.6 &\multicolumn{2}{c}{\makecell[c]{75.48}} &\multicolumn{2}{c}{\makecell[c]{82.62}} &\bf{93.32} &\bf{66.31} &\bf{95.04} &84.69 &66.76 &95.90 &\bf{86.58} &\bf{91.59} &\bf{86.58} &69.39 &79.70 &81.60\\

EIEGSeg(OCR)$^\dag$  &98.44   &\multicolumn{2}{c}{\makecell[c]{87.3}}  &93.54 &\bf{63.27}  &65.81  &\bf{72.3}  &\multicolumn{2}{c}{\makecell[c]{\bf{75.71}}}  &\multicolumn{2}{c}{\makecell[c]{\bf{83.71}}}  &93.08  &65.51  &94.78 & \bf{85.19} & \bf{69.33} &\bf{95.91} &85.15 &88.87 &84.49 &\bf{69.80} & \bf{80.12} &\bf{81.70}\\
\midrule

OCR$^*$  &98.61  &\multicolumn{2}{c}{\makecell[c]{88.11}}  &93.94 &65.67  &67.51 &72.67  &\multicolumn{2}{c}{\makecell[c]{77.00}}  &\multicolumn{2}{c}{\makecell[c]{84.05}}  &\bf{93.58}  &67.75  &\bf{95.11} & 85.66 & 68.44 &\bf{96.24} &\bf{88.05} &\bf{92.73} &\bf{86.13} &71.16 & 80.91 &82.80\\

EIEGSeg(OCR)$^*$$^\dag$  &98.61   &\multicolumn{2}{c}{\makecell[c]{\bf{88.39}}}  &93.94 &\bf{66.71}  &\bf{68.37}  &\bf{73.95}  &\multicolumn{2}{c}{\makecell[c]{\bf{77.67}}}  &\multicolumn{2}{c}{\makecell[c]{\bf{85.06}}}  &93.45  &\bf{69.26}  &95.00 & \bf{86.38} & \bf{71.12} &96.20 &81.36 &91.61 &84.99 &\bf{71.53} & \bf{81.39} &\bf{82.90}\\

\bottomrule

\end{tabular*}}
\end{center}

\end{table*}

As most of the autonomous driving scenarios require real-time performance, we firstly chose two popular lighweight CNN architectures ERFNet and BiSeNetV1-R18 (R18 is ResNet18) as backbones, whose inference speed is 83 and 65.5 frames per second (FPS) on GPU, respectively~\cite{romera2017erfnet,yu2018bisenet}, see Tab.~\ref{tab.city.realtime}. It is noticed that our loss achieves a significant improvement of mIoU on baseline ERFNet (after 150 epoch of training until convergence), up to ${3.19\%}$. BiSeNetV2 ~\cite{yu2021bisenet} is an improved version of BiSeNetV1 with some auxiliary heads on training stage to gain better feature representation ability. Its large version, BiSeNetV2-L has a slower inference speed of FPS 47.3 but lower mIoU of 75.80 compared to EIEGSeg(BSV1) (75.82). This shows that our loss function can guide lightweight networks to better extract detailed features and thus get finer results on exquisite-structures. When using multi-scale inference (random scale on \{0.25, 0.5, 0.75, 1.0, 1.25, 1.5, 1.75, 2.0\}), EIEGSeg achieves 79.0 on backbone BiSeNetV1 (78.86 originally), which outperforms many other models.

In addition, our methods works well on larger size model, such as OCR, a high-resolution non real-time network structure with dense connections and of 0.15 FPS ~\cite{seong2021semantic}. We use the pre-trained model provided by the author and add EIEL to continue the training. In our implementation, our method outperforms the pre-trained model, especially on fine-scale structures. Some other non real-time approaches are compared in the last six rows, while the single-scale and multi-scale test results are displayed in last for rows of \ref{tab.city.realtime}.

Figure \ref{fig.seg} shows semantic segmentation performance on cityscapes. The improved areas trained with EIEL are framed in different colors (the forth column). In the row 2 of right most column, the segmentation of the pedestrian legs is even better than what was provided in ground truth, because EIEGSeg correctly predicts the man in the picture wearing shorts with a gap between his legs.

\subsubsection{Segmentation-based lane detection.}
Figure \ref{fig.process} shows the training process with/without EIEL. The image and down-scaled ground truth are shown in Fig.~\ref{fig2}. As our size of ground truth is $88*160$, the prediction boundaries are also grid-like. The first column is the model outputs after 1st epoch, which are both disconnected on the left most lane and blurry on the right most lane, because they are occluded by cars. Adding EIEL fixes the wrong prediction and finally the lanes are all continuous after convergence. In addition, EIEL makes the prediction much straighter, sharper and precise on details than those without EIEL.

  

\begin{table*}[htbp]\small
\begin{center}
\caption{Comparison of class IoU and total IoU (threshold = 0.5) on baseline ERFNet with / without EIEL on CULane.}\label{tab.cu}
\begin{tabular*}{\hsize}{@{}@{\extracolsep{\fill}}c|ccccccccc|l@{}}
\toprule
Method  &Normal  &Crowded  &Dazzle  &Shadow  &No line   &Arrow   &Curve  &Cross  &Night  &Total \\
\midrule  
ERF~\cite{romera2017erfnet}  &85.60   &67.14  &59.40  &60.72  &44.07  &78.99  &\bf{60.79}  &\bf{1861}  &61.61 &68.82 \\
EIEGSeg(ERF)   &\bf{88.96}   &\bf{71.29}   &\bf{62.66}   &\bf{72.69}   &\bf{47.26}   &\bf{84.14}   &60.55   &2795   &\bf{66.41} &\bf{72.19} \\
\midrule
EIEGSeg(ERF)+IoU+LE   &90.08  &72.21   &63.58   &72.96   &48.95   &85.58   &62.46  &2201   &68.50 &\bf{73.76} \\
\bottomrule
\end{tabular*}
    
\end{center}
\end{table*}

\begin{table}[htbp] 
\caption{Comparison of two baseline with / without EIEL on TuSimple. The batch size is set to be 8 and the training epoch is set 20.}
  \label{tab.tu}
 \begin{center}
 \begin{threeparttable}
  \begin{tabular}{cccccc} 
  \toprule 
   \makecell[c]{Model}&Acc (\%) & FP & FN &F1 & Pix.F1\\
   \midrule 
   ERFNet~\cite{romera2017erfnet}  & 94.91  &    0.061 &     0.069 & 93.52 &98.04\\
   EIEGSeg(ERFNet) & \bf{95.16}  &    \bf{0.061} &     \bf{0.062} & \bf{93.81} &\bf{98.30}\\
  \bottomrule 
  \end{tabular}
  \small

 \end{threeparttable}
  \end{center}
\end{table}

It is noticed that in Fig.~\ref{fig.lane}, our approach can make the shape at the end of the driveway clearer, and improve the mis-classification problem and lane blurry/missing situation. As marked in the pictures in the images, the far lane shape in the first row of images is curved, and the blocked lane line in the third row of image is straight, and our inference accuracy is higher. The disconnection due to incorrect classification in the second row, and the missing lane lines due to occlusion in the fourth and fifth rows are also fixed. The experiments shows that EIEL can indeed improve the details segmentation, and have better inference ability in poor visual conditions.



Table \ref{tab.cu} and \ref{tab.tu} shows the results on lane predictions of CULane and Tusimple, respectively. In CULane results, most of the category IoU increase with the assistance of EIEL as displayed in first two rows of Tab.~\ref{tab.cu}. Last row in Tab.\ref{tab.cu} is the results when adding two auxiliary loss functions $L_{IoU}$ and $L_{exist}$, where the former one guides the network to increase the pixel interaction of union, and the latter one is achieved by add an auxiliary branch of lane existence (LE), and calculated by binary cross entropy~\cite{pan2018spatial}. In experiment results of Tusimple, EIEL makes consistently improvement on every metrics shown in Tab.~\ref{tab.tu}. These results show that the use of EIEL can significantly improve lane segmentation results and is suitable for any task that requires to obtain lane heat maps/probability maps. There are many other ways to tune the parameters, including  setting the tusimple labels (5 lanes from left to fight, or 4 / 6 lanes on two sides of ego lane), batch size, learning rates, adding various auxiliary branches, etc. We do not focus too much on these improvement measures, but just make comparison on segmentation quality via experiments on the same code and machine until convergence. 

\begin{figure*}[htbp]
\centering
\includegraphics[width=1.0\textwidth]{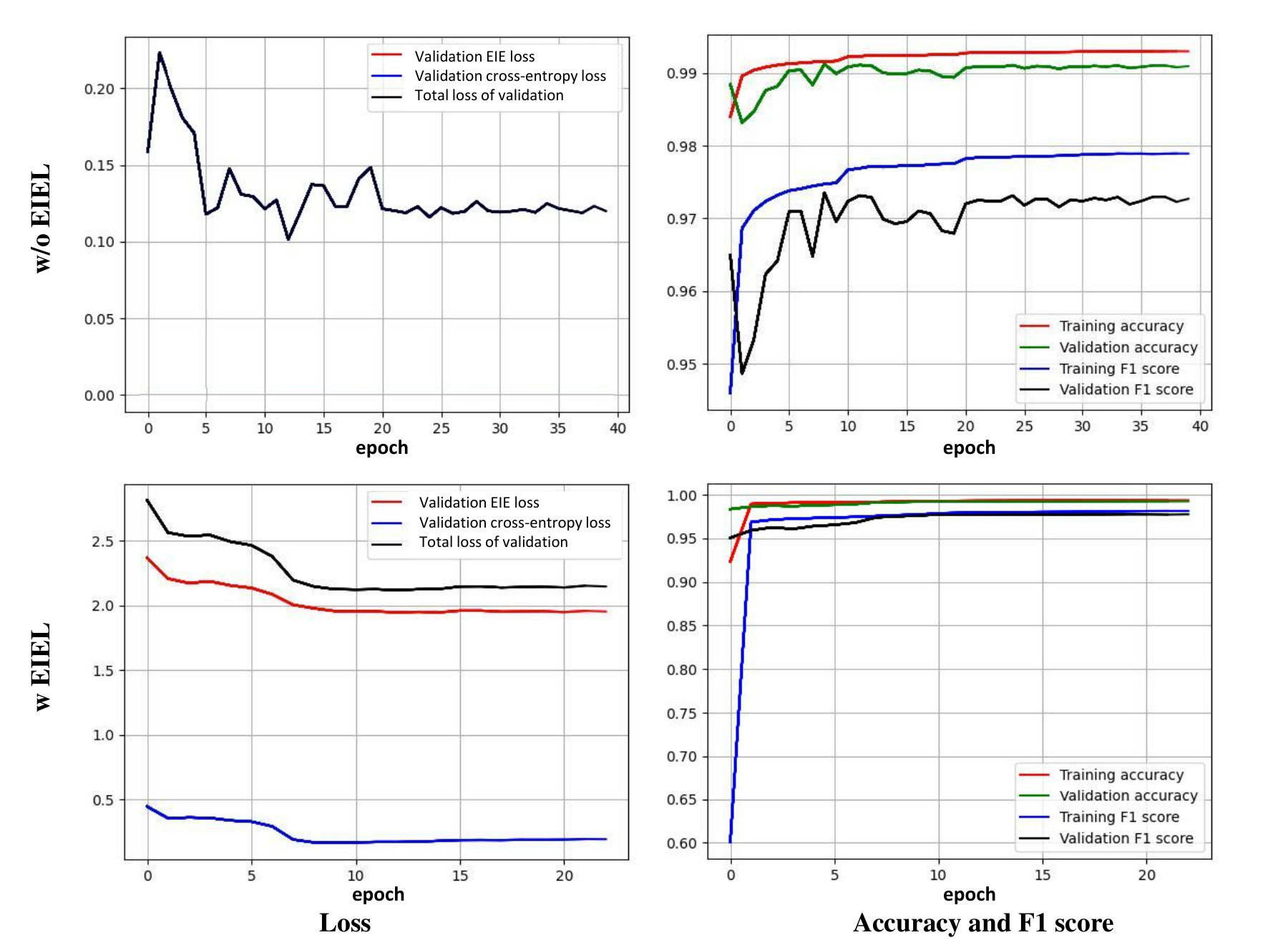} 
\caption{Illustration on how EIEL jointly trained with CE influences the loss value and pixel accuracy. EIEGSeg training strategy converges faster.}
\label{fig.curve}
\end{figure*}

Figure \ref{fig.curve} shows some curves of the training process. The left column are validation losses, and the right column are average pixel accuracy and F1 measure. (a) and (b) are the experimental diagrams without EIEL, and (c) and (d) are the experimental diagrams with EIEL. Our EIEGSeg achieves early convergence (from about epoch 10), while the baseline curve oscillations do not flatten out until epoch 20. Due to the long-range nature of our loss, the model EIEGSeg, is insensitive to the initial distribution and has the properties of training stability and rapid convergence.

\section{CONCLUSION}
We proposed an EIEGSeg training strategy for multi class segmentation task in autonomous driving. It is proved to be robust on segmentation of geometry complex objects such as long-thin, small or irregular shaped objects. It can also be easily combined into different network architectures, especially the light-wight backbones, which can help achieve real-time inference in autonomous driving.

\section*{Acknowledgement}
This work is supported by the Project of Hetao Shenzhen-HKUST Innovation Cooperation Zone HZQBKCZYB-2020083, the project Study on Algorithms for Traffic Target Recognition in Autonomous Driving at HKUST Shenzhen Research Institute and Shenzhen Science and Technology Program (No. KQTD20180411143338837).

\bibliographystyle{unsrt}  
\bibliography{references}  


\end{document}